\documentclass[letterpaper]{article} 
\usepackage{aaai2026}  
\usepackage{times}  
\usepackage{helvet}  
\usepackage{courier}  
\usepackage[hyphens]{url}  
\usepackage{graphicx} 
\usepackage{natbib}  
\usepackage{caption} 
\usepackage{multirow}
\usepackage{booktabs}
\usepackage{graphicx}
\usepackage{algorithm}
\usepackage{algpseudocode}
\usepackage{caption}
\usepackage{subfigure}
\usepackage{amssymb}
\usepackage{soul}

\usepackage[utf8]{inputenc} 
\usepackage[T1]{fontenc}    

\usepackage{url}            
\usepackage{booktabs}       
\usepackage{amsfonts}       
\usepackage{nicefrac}       
\usepackage{microtype}      
\usepackage{xcolor}
\usepackage{amsmath}
\usepackage{amssymb}
\usepackage{pgf-pie}
\usepackage{graphicx}
\usepackage{pgfplots}
\usetikzlibrary{patterns}
\pgfplotsset{compat=1.18}
\usepackage{graphicx}
\usepackage{pythonhighlight}
\usepackage{bbding}
\usepackage{listings}

\usepackage{array}
\usepackage{fvextra}
\usepackage{listings}

\usepackage{tikz}
\usepackage{fontawesome5}
\usepackage{makecell}
\usepackage{tcolorbox}
\tcbset{
  colframe=black!15, colback=gray!5, boxrule=0.8pt, sharp corners,
  boxsep=0.8em, left=0.2em, right=0.2em, top=0.5em, bottom=0.5em,
  fonttitle=\bfseries, before skip=4pt, after skip=4pt
}
\newtcolorbox{userbox}{colframe=blue!50!black, colback=blue!5, title=User}
\newtcolorbox{assistantbox}{colframe=green!50!black, colback=green!7, title=Assistant}
\newtcolorbox{thinkbox}{colframe=orange!80!black, colback=orange!8, title=Assistant (Internal Reasoning)}
\newtcolorbox{toolbox}{colframe=black!60, colback=gray!10, title=Assistant (Tool Call/Observation)}
\newtcolorbox{rewardbox}{colframe=purple!70!black, colback=purple!5, title=Ground Truth}

\usepackage{newfloat}
\usepackage{listings}
\DeclareCaptionStyle{ruled}{labelfont=normalfont,labelsep=colon,strut=off} 
\floatstyle{ruled}
\newfloat{listing}{tb}{lst}{}
\floatname{listing}{Listing}
\newcommand\model{ShoppingBench}
\newcommand\product{Products Finder}
\newcommand\store{Multi-products seller}
\newcommand\voucher{Coupon \& Budget }
\newcommand\knowledge{Knowledge}
\newcommand\matchscore{CAR}
\newcommand\successrate{ASR}

\title{ \model: A Real-World Intent-Grounded Shopping Benchmark for LLM-based Agents}
\author{
    Jiangyuan Wang \textsuperscript{\rm 1}\equalcontrib, Kejun Xiao\textsuperscript{\rm 1}\equalcontrib, Qi Sun\textsuperscript{\rm 1}\thanks{Corresponding author.}, \\Huaipeng Zhao\textsuperscript{\rm 1}, Tao Luo\textsuperscript{\rm 1}, Jian Dong Zhang\textsuperscript{\rm 1}, Xiaoyi Zeng\textsuperscript{\rm 1} \\
}
\affiliations{
    \textsuperscript{\rm 1}Alibaba International Digital Commercial Group\\

    \{wangjiangyuan.wjy, xiaokejunkejun.xia, qiran.sq\}@alibaba-inc.com\\
    
}

\usepackage{bibentry}

\begin{document}

\maketitle

\begin{abstract}

Existing benchmarks in e-commerce primarily focus on basic user intents, such as finding or purchasing products. However, real-world users often pursue more complex goals, such as applying vouchers, managing budgets, and finding multi-products seller. To bridge this gap, we propose \model{}, a novel end-to-end shopping benchmark designed to encompass increasingly challenging levels of grounded intent. Specifically, we propose a scalable framework to simulate user instructions based on various intents derived from sampled real-world products. To facilitate consistent and reliable evaluations, we provide a large-scale shopping sandbox that serves as an interactive simulated environment, incorporating over 2.5 million real-world products. Experimental results demonstrate that even state-of-the-art language agents (such as GPT-4.1) achieve absolute success rates under 50\% on our benchmark tasks, highlighting the significant challenges posed by our \model{}. In addition, we propose a trajectory distillation strategy and leverage supervised fine-tuning, along with reinforcement learning on synthetic trajectories, to distill the capabilities of a large language agent into a smaller one. As a result, our trained agent achieves competitive performance compared to GPT-4.1\footnote{Released on \url{https://github.com/yjwjy/ShoppingBench}}.

\end{abstract}

\section{Introduction}

Large language models (LLMs) have empowered agents with impressive abilities in task automation and decision-making, leading to growing interest from both academia and industry \cite{tau_bench}. In recent years, a variety of agent benchmarks have been introduced to systematically assess language agent performance across different scenarios. These benchmarks typically focus on evaluating end-to-end capabilities such as task planning, tool using, and reasoning. As a highly practical field with broad application prospects, e-commerce has naturally become a key focus for evaluating agent capabilities. 

\begin{figure*}
	\centering
	\includegraphics[width=0.65\textheight]{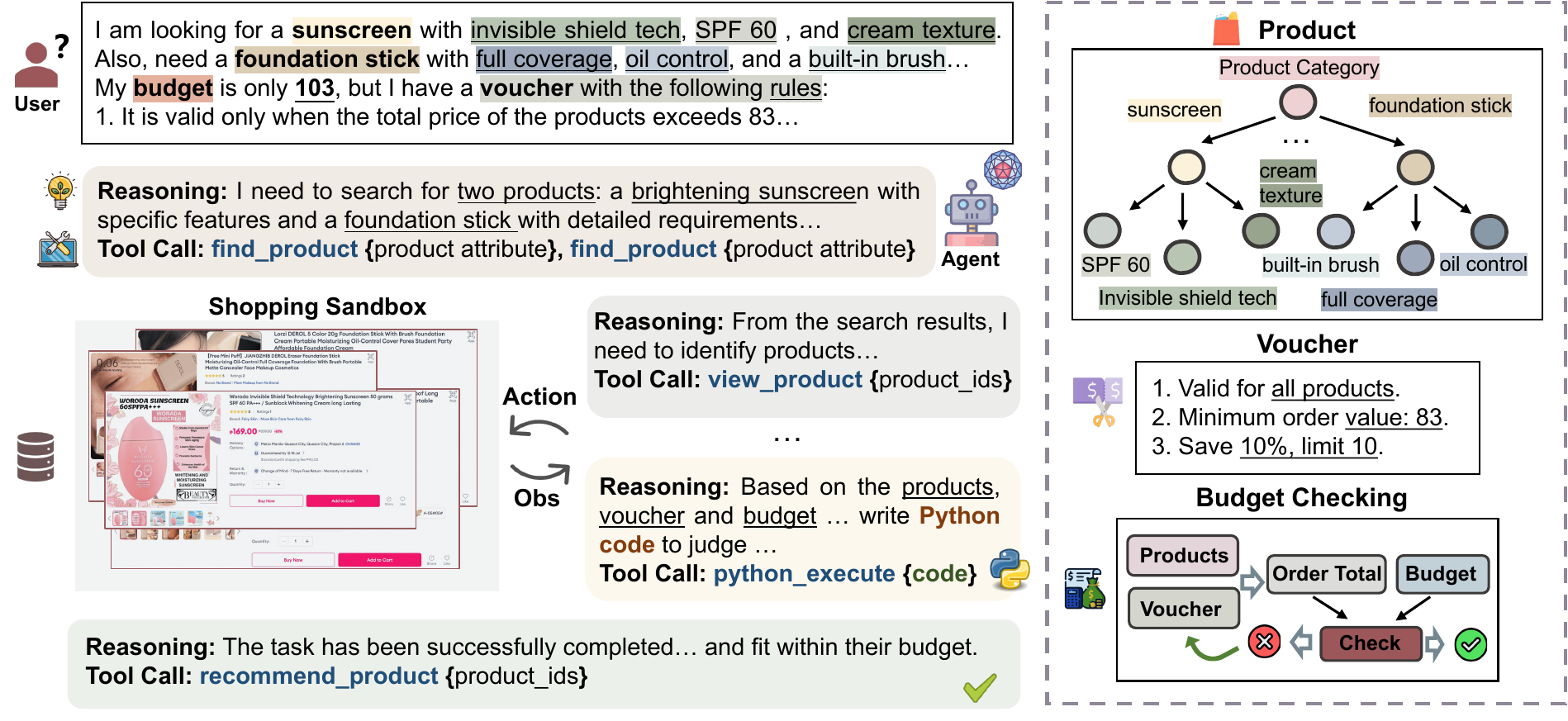}
	\caption{An illustration to depict a real-world user instruction with complex intent. Unlike previous agent benchmarks that solely focus on basic product purchases, this example incorporates coupon usage and optimal product combination within a budget.}
 \label{fig:intro}
 
\end{figure*}

However, existing benchmarks for evaluating language agents in e-commerce primarily focus on straightforward user intents such as locating and purchasing products \cite{WebArena,Webshop}. In practice, e-commerce users often pursue multifaceted goals that extend beyond mere product acquisition. For example, as shown in Figure \ref{fig:intro}, language agent is expected to optimize discounts, combine multiple orders to qualify for free shipping, or verify total expenditures against budget constraints. Such grounded user intents require language agents to perform multi-step reasoning, effectively utilize domain-specific knowledge, and leverage external tools to fulfill complex user instructions. Despite increasing interest in language agents as autonomous decision-makers \cite{Gaia}, current agent benchmarks in e-commerce rarely incorporate these realistic and nuanced user intents.

Beyond above agent benchmarks, previous e-commerce datasets primarily address isolated or narrowly scoped downstream tasks \cite{Amazon-M2,Shopping_Queries_Dataset,EcomGPT,liu2023u,jia2022convrec}. While large-scale benchmarks such as Shopping MMLU \cite{Shopping_MMLU} and ChineseEcomQA \cite{Chineseecomqa} have been proposed based on large e-commerce corpora, they mainly focus on question answering\cite{EcomScriptBench} and skill-based evaluation rather than end-to-end agent performance. This limits their effectiveness in assessing a language agent's ability to fulfill complex user intents in real-world shopping scenarios.

To bridge the above gaps, we propose \model{}, a large-scale end-to-end shopping benchmark comprising 3,310 user instructions, designed to encompass progressively challenging levels of grounded intent in shopping scenarios. Specifically, we propose a scalable framework to simulate user instructions based on various intents derived from sampled real-world products. To facilitate consistent and reliable evaluations, we provide a large-scale e-commerce shopping sandbox that serves as an interactive simulated environment, incorporating over 2.5 million real-world products. To automatically evaluate the quality of language agents, we propose a series of new metrics based on different intent constraints. 

In addition, we also propose a trajectory distillation strategy, where tool-use trajectories are generated by the GPT-4.1, and using rejection sampling to filter low-quality trajectories. Then, we use these synthetic trajectories to train Qwen3-4B with Supervised Fine-Tuning (SFT) and Reinforcement Learning (RL), which can significantly improve the performance. As a result, our trained language agent achieves competitive performance compared to GPT-4.1 agent. 

Our experiments illustrate that even the best-performing language agent (GPT-4.1-based) achieves a success rate below 50\%, underscoring the challenge of our benchmark. Quantitative and qualitative analysis of failure cases reveals existing agents' limitations in understanding user instruction with complex intent and choosing appropriate tools. These findings underscore the need for advances in agent architecture, tool usage, problem decomposition, and web information integration.

Our contribution can be summarized as follows: 

\begin{itemize}
    
    \item  We propose a scalable framework to simulate diverse user instructions and provide a sandbox environment with over 2.5 million products for consistent and interactive evaluation.
    
    \item We propose new automatic evaluation metrics, grounded in intent constraints, to rigorously assess language agents in e-commerce shopping tasks.

    \item We propose a trajectory distillation strategy to synthesize training data, filter out low-quality trajectories using our proposed automatic evaluation metric, and then use SFT and RL to efficiently distill GPT-4.1's abilities into a smaller model, which achieves comparable performance.

    \item We evaluate 17 existing language agents, along with our fine-tuned Qwen3-4B agent. Even the best-performing model, GPT-4.1, achieves a success rate below 50\%, highlighting the challenge of our benchmark.
    
\end{itemize}

\section{Related Work}
Existing benchmarks related to e-commerce shopping can generally be categorized into two types: agent benchmarks and task-oriented dialogue benchmarks. 

\subsection{Agent Benchmarks}

Recent advances in language agents have generated significant interest regarding their potential to drive unprecedented automation across diverse industries \cite{chen2021evaluating,ReAct}. Evaluating the capabilities of language models as agents requires examining their ability to aggregate information for multi-step reasoning and autonomous decision-making, as well as their proficiency in effective tool utilization \cite{MetaTool,schick2023toolformer,tau_bench}. Recent research focus on developing domain-specific agents \cite{Gaia} to address these challenges. E-commerce represents an especially realistic and pressing application scenario for language agents. Thus, more recent studies adopted web shopping \cite{WebArena, Webshop} as a key benchmark domain to evaluate the capabilities of language agents in fulfilling user purchase requests. Most existing benchmarks for agents in e-commerce scenarios primarily focus on evaluating the user's basic intent, namely, the successful purchase of a product. These benchmarks typically adopt a web shopping setting and define task success based on whether an order can be successfully placed.

However, in real-world e-commerce scenarios, user intents often extend beyond the basic goal of finding products. More complex and realistic intents, such as combining orders for free shipping or optimizing for coupon discounts, require language models to reason with specific e-commerce knowledge that are not adequate in existing agent benchmarks. To bridge this gap, we propose \model{}, an end-to-end shopping benchmark for language agent, which grounds a wide range of realistic user intents.

\begin{figure*}[h]
	\centering
    \includegraphics[width=0.7\textheight]{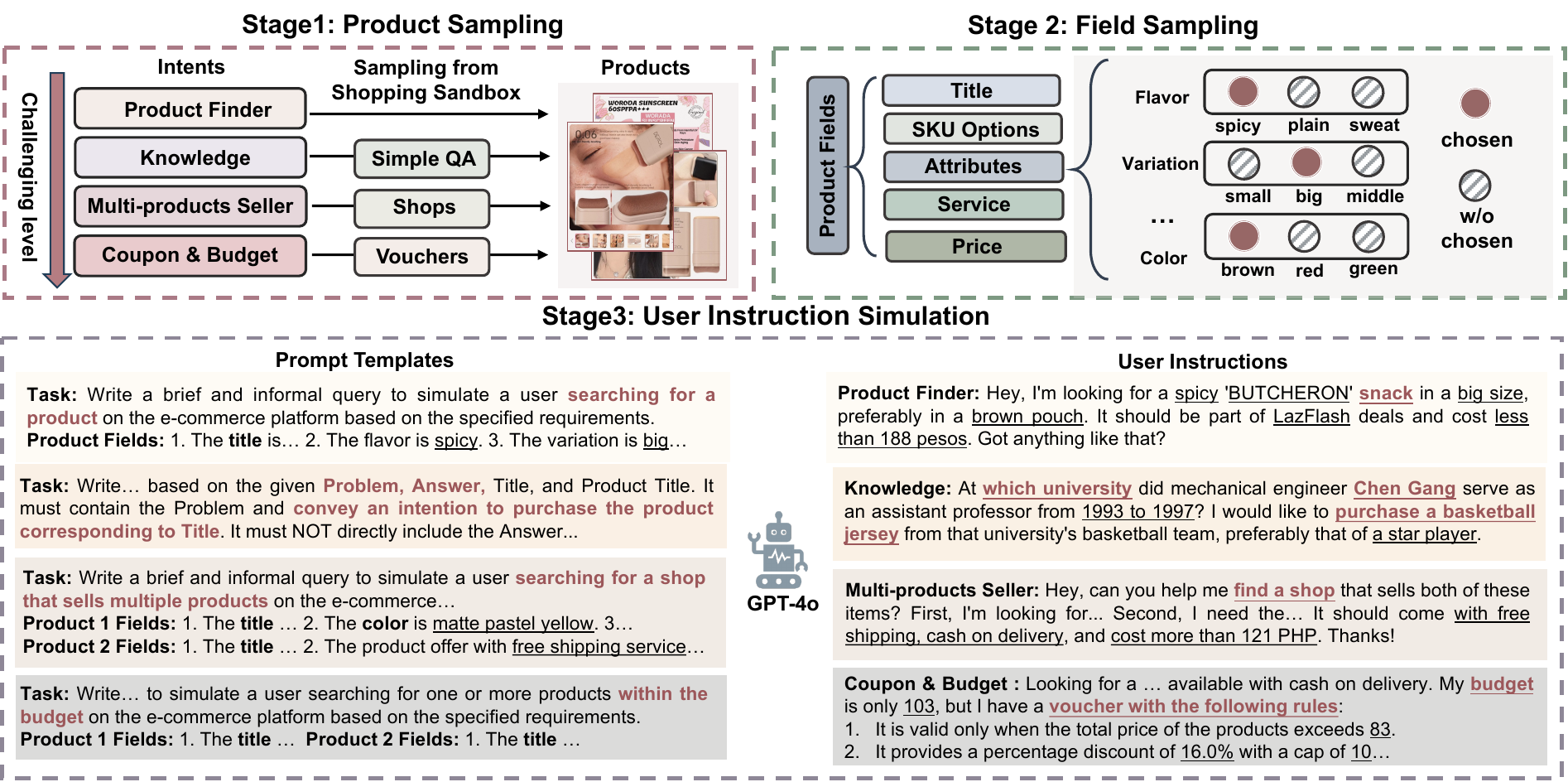}
	\caption{Construction of our \model{}. Our benchmark encompasses four types of real-world user purchase intents: \product{}, \knowledge{}, \store{}, and \voucher{}, with complexity increasing progressively. }
 
 \label{fig:pipeline}
\end{figure*}

\subsection{E-commerce Datasets}
Previous e-commerce benchmarks mainly comprise isolated or narrowly related downstream tasks \cite{Amazon-M2,Shopping_Queries_Dataset,EcomGPT,liu2023u,jia2022convrec}. Recently, multidimensional benchmarks such as Shopping MMLU \cite{Shopping_MMLU} and ChineseEcomQA \cite{Chineseecomqa} have been constructed based on comprehensive e-commerce corpora. EcomScriptBench \cite{EcomScriptBench} is proposed to evaluate the ability of language models to generate plans with scripts and recommend products. However, these benchmarks primarily focus on conceptual and skill-based question answering in e-commerce, which poses challenges for the end-to-end evaluation of e-commerce agents.

\section{Problem Formulation}
Each trajectory can be represented as a partially observable Markov decision process (POMDP) $(\mathcal{U}, \mathcal{S}, \mathcal{A},\mathcal{T},  \mathcal{O}, \mathcal{R})$. It consists of the following components: a natural language instruction space $\mathcal{U}$, a state space $\mathcal{S}$, an action space $\mathcal{A}$, an observation space $\mathcal{O}$, a transition function $\mathcal{T}$, and a reward function $\mathcal{R}$. When a shopping agent performs an action, it interacts with the environment by invoking tools, which generates observations and updates the state. This process can be mathematically represented as:

\[
(s_{t+1}, o_{t+1}) = \mathcal{T}(s_t, a_t)
\]

where \(s_t\) represents the state at time step \(t\), \(a_t\) denotes the action executed by the agent, \(\mathcal{T}\) is the transition function, \(s_{t+1}\) is the updated state, and \(o_{t+1}\) is the observation.

At the terminal state, we evaluate whether the predicted products in the terminal state, \( s_T \), satisfy all the requirements specified in the user's instructions, \( \mathcal{U} \). The task is deemed successfully completed if:

\[
\text{success} = 
\begin{cases} 
1, & \text{if all conditions in } \mathcal{U} \text{ are met in } s_T, \\
0, & \text{otherwise.}
\end{cases}
\]

\section{\model{} Construction}
In this section, we introduce the construction of our \model{}, which consists of three key components: a simulated interactive environment, intent-grounded user instructions, and a predefined tool set.

\subsection{Grounded Shopping Intention}
As shown in Figure \ref{fig:pipeline}, our \model{} includes the following four real-world user purchase intents, with the challenges progressively increasing for each intent.

\paragraph{\product.}The language agent needs to find the corresponding product based on the user's description of the product attributes.

\paragraph{\knowledge.}The language agent needs to infer the knowledge in the user's question and identify the related product.

\paragraph{\store.}The language model needs to find the store that sells all the products described by the user.

\paragraph{\voucher.}The language agent needs to understand the voucher rules and find optimal product combinations within a budget.

\subsection{Shopping Sandbox Implements}
To ensure more consistent and reliable evaluations, we offer a large-scale shopping sandbox, serving as a simulated interactive environment that incorporates over 2.5 million real-world products sourced from Lazada.com. In this environment, an AI shopping agent is tasked with recommending suitable products tailored to the user's real-world intents by leveraging a variety of tools. To support the API tool, we build two search engines: one for the product database and another for web-based knowledge.

\paragraph{Search engine.} We use Pyserini\cite{Lin_etal_SIGIR2021_Pyserini} to build a product search engine, utilizing the BM25 sparse retrieval model to construct the index offline. 
\paragraph{Web Search engine.} We encapsulated a web search tool using Serper\footnote{\url{https://serper.dev/}}, enabling access to online searches.

\subsection{Intent-Grounded Instruction Generation.}
In this subsection, we present the framework for generating intent-grounded user instructions, which encompasses three key stages.

\paragraph{Stage I: Sampling Real-World Products}
We begin by sampling a diverse set of real-world products from our shopping sandbox, ensuring coverage across a wide range of constraints such as variations in categories, brands, attributes, and service. The sampling distribution of the product can be seen in Appendix A. Notably, for the \knowledge{} intent, SimpleQA \cite{simpleQA} is used to link products, ensuring the verifiability of the responses. For the \voucher{} intent, we also synthesize multiple voucher rules and sample products that meet these rule requirements.

\paragraph{Stage II: Extracting Product Fields}
In the second stage, specific fields are extracted from the sampled products. These fields include detailed information such as product titles, attributes, associated services, and other relevant meta data. This structured information forms the basis for developing realistic user scenarios.

\paragraph{Stage III: Simulating User Queries}
Using the extracted product fields, we employ GPT-4.1 to simulate diverse and realistic user queries. These queries are carefully tailored to align with each purchase intent, which can be seen in Figure \ref{fig:pipeline}. Each simulated user instruction is intent-specific and grounded in real-world scenarios, aiming to evaluate the model's capability to understand and navigate the constraints inherent in e-commerce tasks.

\begin{figure}
	\centering
	\includegraphics[width=0.33\textheight]{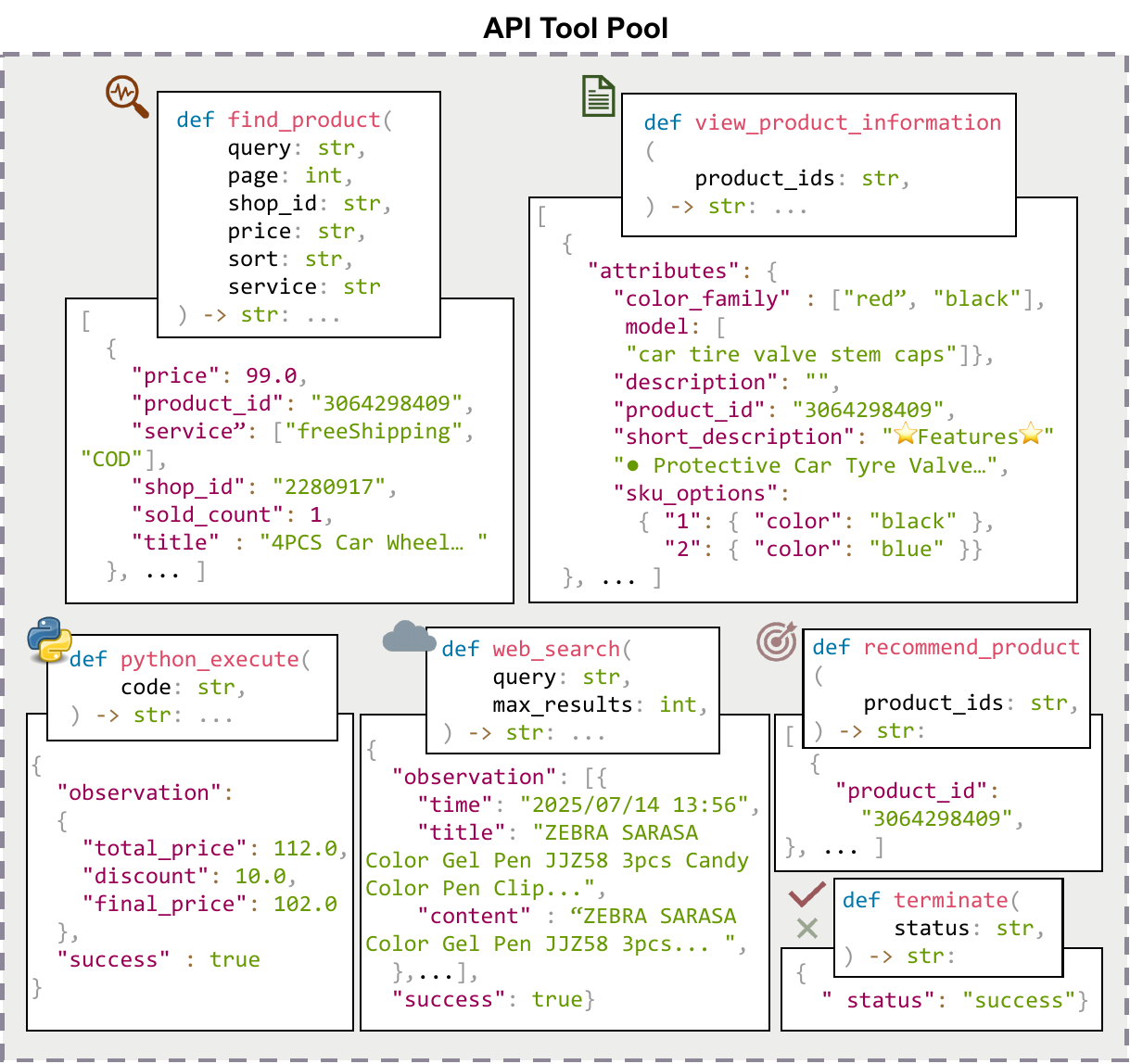}
    
	\caption{\model{} provides six API tools designed to facilitate agent interaction with our shopping sandbox environment. }

 \label{fig:tool}
\end{figure}

\subsection{Interaction Tools}
We provide a set of API tools to interact with our shopping sandbox. As shown in the Figure \ref{fig:tool}, we design six API tools: retrieving product lists, viewing product details, calculating discounts and budgets, retrieving web knowledge, recommending products, and terminating states. Each invocation of a tool is formalized as an action \(a_t\), specifically represented as \(\text{tool\_name}(para)\), where \(\text{tool\_name}\) denotes the name of the tool being called, and \(para\) represents the parameters passed to the tool. Upon invocation, the tool returns an observation \(o_{t+1}\), which is the result or output of tool execution.

\section{Evaluation}
\label{sec:metric}

Given a user instruction, the model outputs its reasoning process as well as an action in the form of a tool call in each step\cite{ReAct}. Based on the predicted tool calling, we parse the list of invoked tools and execute the corresponding functions, returning the observation. Then the agent generates the next round of reasoning and action predictions based on the current observation and the user instruction. This process is repeated until a terminal tool is called to end the current trajectory. In the terminal state, we compare the predicted products with the target products to automatically determine whether the task has been successfully completed.

\subsection{Constrain Scores}
We design the Absolute Success Rate (\successrate{}) and Cumulative Average of the product Relevance (\matchscore{}) as metrics, which calculated from the following constrain scores.

\paragraph{Product Relevance Score.} 
To measure the relevance between the predicted product and the target product, we consider three dimensions: title similarity, price similarity, and product feature similarity. The formulation can be seen as follows:

\begin{equation}
    r_{\text{pro}} = \frac{\mathbb{I}_{sim \geq 0.5}+\mathbb{I}_{min \leq p \leq max}+|F_{\text{t}} \cap F_{\text{p}}|}{2+|F_{\text{t}}|}, 
\label{eq:product_score}
\end{equation}

where $\mathbb{I}_{sim \geq 0.5}$ indicates that when the title similarity between the predicted product and the target product exceeds the threshold (set to 0.5), the value is 1. $\mathbb{I}_{min \leq p \leq max}$indicates that when the price \(p\) is within the target product price range \([min, max]\), the value is 1. \(|F_{\text{t}} \cap F_{\text{p}}|\) is the number of overlapping features, and \(|F_{\text{t}}|\) is the total number of features in the target product. 

\paragraph{Knowledge Constrain Score.}
For the \knowledge{} intent, we evaluate whether the predicted products have the correct knowledge attribute, as follows:

\begin{equation}
r_{\text{kw}} = 
\begin{cases} 
1, & \text{if } \text{knowledge\_attribute in title}, \\
0, & \text{otherwise}.
\end{cases}
\label{eq:knowledge}
\end{equation}

\paragraph{Shop Constrain Score.} 

For \store{} intents, to assess whether the predicted products satisfy the user's request that all products come from the same shop, we define the shop relevance score as follows: 

\begin{equation}
r_{\text{shop}} = 
\begin{cases} 
1, & n_{\text{t}} = \ n_\text{p} \, \text{and}\,  |S_{\text{rec}}| = 1 \\
0, & \text{otherwise}
\end{cases}
\label{eq:shop_relevece}
\end{equation}

The score is 1 when the number of predicted products $n_p$ is equal to the number of target products $n_t$ and all predicted products come from the same store ($|S_{\text{rec}}| = 1$).  

\paragraph{Budget Constrain Score.}

For \voucher{} intent, to evaluate whether the predicted products meet the user's budget , we define the budget score as follows:

\begin{equation}
r_{\text{budget}} = 
\begin{cases} 
1, & \text{if } \text{total\_price} \leq \text{budget}, \\
0, & \text{otherwise}.
\end{cases}
\label{eq:budget}
\end{equation}

\subsection{Overall Metrics}
In this subsection, we detailed introduce our metrics: the Absolute Success Rate (\successrate{}) and the Cumulative Average of the product Relevance (\matchscore{}).

\paragraph{Cumulative Average of the Product Relevance}

For each intention, we compute the Cumulative Average of the product Relevance (\matchscore{}) between the predicted and target products, defined as:

\begin{equation}
A_{\text{pro}} = \frac{1}{n} \sum_{i=1}^{n} \frac{1}{n_i} \sum_{j=1}^{n_i} r_{\text{pro}}^{(j)},
\label{eq:match_score}
\end{equation}

where $n$ indicates the number of samples. \( \frac{1}{n_i} \sum_{j=1}^{n_i} r_{\text{pro}}^{(j)} \) represents the average product relevance within the \(i\)-th sample. $n_i$ indicates the number of products in \(i\)-th sample.

\paragraph{Absolute Success Rate}

We also further design the following metrics to measure the absolute success rate (\successrate{}) for each intent.

\paragraph{\product:} For intents where the user wants to locate a particular product according to its attributes, we use the product relevance score $r_{\text{pro}}$ (Equation \ref{eq:product_score}) to determine task success.

\begin{equation}
S_{\text{pro}} = \frac{1}{n}\sum_{i=1}^{n} \delta(r_{\text{pro}}^{(i)} = 1),
\end{equation}

where, the indicator function \( \delta(\cdot) \) is defined as:

\begin{equation}
\delta(r_{\text{pro}}^{(i)} = 1) =
\begin{cases}
1, & \text{if } r_{\text{pro}}^{(i)} = 1, \\
0, & \text{otherwise}.
\end{cases}
\end{equation}

\paragraph{\knowledge:}For intents where user instructions require knowledge reasoning, we include the knowledge constraint score $r_{\text{kw}}$ (Equation \ref{eq:knowledge}).

\begin{equation}
S_{\text{kw}} = \frac{1}{n} \sum_{i=1}^{n} \delta(r_{\text{pro}}^{(i)} = 1, r_{\text{kw}}^{(i)} = 1).
\end{equation}

\paragraph{\store:} For the intent where users want to find multiple products sold by the same shop, we introduce the shop constraint score $r_{\text{shop}}$ (Equation \ref{eq:shop_relevece}).
\begin{equation}
S_{\text{shop}} = \frac{1}{n} \sum_{i=1}^{n} \delta(\frac{1}{n_i}\sum_{j=1}^{n_i}r_{\text{pro}}^{(j)} = 1, r_{\text{shop}}^{(i)} = 1).
\end{equation}

\paragraph{\voucher:} For the intent with budget requirements, we introduce the budget constraint score $r_{\text{budget}}$ (Equation \ref{eq:budget}). The formulation is defined as follows:

\begin{equation}
S_{\text{budget}} = \frac{1}{n} \sum_{i=1}^{n} \delta(\frac{1}{n_i}\sum_{j=1}^{n_i}r_{\text{pro}}^{(j)} = 1, r_{\text{budget}}^{(i)} = 1).
\end{equation}

\section{Shopping Agent Training}

We utilize synthetic trajectories to train Qwen3-4B backbone using Supervised Fine-Tuning (SFT)  and tool-calling based Reinforced Learning (RL).

\paragraph{Trajectory Distillation.} We leverage GPT-4.1 to generate tool-calling trajectories from 2,410 user instructions. According to our evaluation metrics, we apply rejection sampling to filter out low-quality trajectories, specifically excluding those that are not absolute successes.

\paragraph{Cold Start with SFT.} We sample multiple steps from each trajectory. The final training dataset includes 5,552 steps. The model input includes the user instruction as well as the observation (e.g., a retrieved product list). The output consists of a reasoning trace (the process) and the next action (tool-calling). Then, we perform SFT on Qwen3-4B to enhance the model's ability to understand complex instructions, process multi-round observations, and predict actions.

\paragraph{Reinforced Tool Calling.}To further enhance the model's tool-calling capabilities, we apply GRPO \cite{grpo} with the tool reward\cite{toolrl} to continue training the SFT-Qwen3-4B model. The reward function consists of a format reward and a tool reward. The tool reward is calculated based on the matching score between the predicted and target tool's name, parameters, and values.

\begin{table*}[h]
\setlength\tabcolsep{5 pt}
\centering
\resizebox{\linewidth}{!}{
\begin{tabular}{l c c c c c c c c c  }
\toprule
\multirow{2}*{\textbf{Models}}& \multicolumn{2}{c}{\textbf{\product}}& \multicolumn{2}{c}{\textbf{\knowledge}} & \multicolumn{2}{c}{\textbf{\store}} & \multicolumn{2}{c}{\textbf{\voucher}}  & \multirow{2}*{\textbf{Avg.}}\\
\cmidrule(r){2-3}
\cmidrule(r){4-5}
\cmidrule(r){6-7}
\cmidrule(r){8-9}
 &  \successrate{}(\%) & \matchscore{}(\%) & \successrate{}(\%) & \matchscore{}(\%) & \successrate{}(\%) & \matchscore{}(\%) & \successrate{}(\%) & \matchscore{}(\%)  \\
\midrule
\multicolumn{10}{c}{\textbf{Closed-Source Large Language Models}} \\
\midrule
GPT-4.1  & \underline{59.6}	& \underline{83.6}	& \underline{\textbf{62.0}}	& \underline{\textbf{67.3}}	& \underline{46.4}	& \underline{79.2}	&30.4&	\underline{72.8} & \underline{48.2}\\
o3-mini &42.0&	62.6&	51.3	&57.3&	36.8&	50.3	& \underline{31.6} &	61.4 &39.2\\
GPT-4o &52.4	&71.5&	50.0&	58.7&	24.0&	52.4&	25.2	&65.6&36.6\\
GPT-4o-mini  &33.2	&46.9	&28.0&	31.3	&10.4&	52.7&	11.6&	54.6 &20.0\\
Gemini-2.5-Flash &49.2	&71.3&	39.3&	46.7&	32.0&	40.9&	22.8&	55.0&35.4\\
Claude-4-Sonnet &48.0&	73.1&	51.3&	62.7 &37.6	&59.5 &24.0	&72.2&39.0\\

Qwen2.5-max &58.4&	81.0&	42.7&	50.7	&22.8	&67.3&	22.4&	65.8 &35.9 \\
\midrule
\multicolumn{10}{c}{\textbf{Open-Source Large Language Models}} \\
\midrule
DeepSeek-R1  &53.2&	75.8&	44.0	&53.3&	\underline{37.2}	& 51.5 & \underline{24.4} &	43.9& \underline{39.2} \\
DeepSeek-V3 & \underline{54.8}	&75.8&	\underline{48.0} &	54.7 &	22.8&	46.9&	21.2&	55.3& 35.4\\
Qwen3-235B-A22B &49.2&	77.2&	40.0&	46.7&	28.8&	56.3&	14.4&	55.3&	32.3\\
Qwen3-32B &51.6&	\underline{77.6} &	45.3&	54.0&	25.6	&54.2&	18.0&	\underline{63.8}	&34.0\\
Qwen3-14B &46.0&	70.4&	30.0&	37.3&	19.2&	53.1&	12.4&	58.1	&26.6\\
Qwen3-8B &40.0	&65.8&	22.7&	27.3&	13.6&	31.7&	11.2&	53.2 & 21.8 \\
Qwen3-4B  & 36.4&	66.4&	18.7&	26.0	&8.8&	29.8&	8.4&	45.5	&18.0\\
Gemma-3-27B &32.0	&48.5& 46.7	& \underline{57.3} &18.0	&\underline{65.4} &17.2	&62.5 &26.5\\
Gemma-3-12B &27.2	&42.5 &32.0&	36.7& 9.6	&51.7 &13.6	&55.7 &19.3\\
Gemma-3-4B&24.4	&40.1 &16.7&	20.7& 0	&31.1 &4.8	&30.7	&10.9\\

\midrule
\multicolumn{10}{c}{\textbf{Ours}} \\
\midrule
SFT-Qwen3-4B & 55.6	&81.1 & \underline{52.7} &	\underline{59.3} & 39.2&	77.9 &30.4	&76.0 &43.6\\
SFT+RL-Qwen3-4B &\underline{\textbf{60.8}}&	\underline{\textbf{86.1}} & 46.7&	51.3& \underline{\textbf{53.2}} &	\underline{\textbf{85.5}} & \underline{\textbf{33.2}} &	\underline{\textbf{79.0}} & \underline{\textbf{48.7}} \\

\bottomrule

\end{tabular}
}
 
\caption{\label{tab:main_result}
Main results of different language agents on our \model{}, including absolute success rate (\successrate) and cumulative average of the product relevance (\matchscore). The average reported is domain-weighted \successrate, rather than task-weighted.
}

\end{table*}

\section{Experiments}

\paragraph{Baselines.}We evaluate 17 language agents on our \model{}, including leading closed-source models (e.g., GPT-4.1, Claude-4-Sonnet, Qwen2.5-max) and open-source models (e.g., DeepSeek-R1, Qwen3-32B, Gemma-3-27B). Additionally, we further train Qwen3-4B with synthetic tool-use trajectories using supervised fine-tuning and reinforcement learning. The training details can be seen in Appendix B.

\paragraph{Dataset.}The \model{} dataset consists of 3,310 user instructions in total, with 2,410 used for training and 900 for testing. The test set includes 150 samples for the \knowledge{} intent and 250 samples for each of the other intents. Detailed data statistics can be seen in Appendix C.

\subsection{Main Results}
As shown in Table~\ref{tab:main_result}, we evaluated various language agents, including both open-source and proprietary ones. The experimental results lead to the following conclusions:

\begin{itemize}
    \item \textbf{Overall Performance:} On untrained language agents, GPT-4.1 achieves the highest overall performance, with an Absolute Success Rate (\successrate) of 48.2\%. Among open-source models, DeepSeek-R1 achieves the strongest overall performance, surpassing GPT-4o in average benchmarks. For simple intents such as product finding, GPT-4.1 reaches 59.6\% \successrate, with cumulative average of product relevance (\matchscore{}) up to 83.6\%. However, the performance of GPT-4.1 drops significantly on complex tasks such as the \voucher{} intent, falling to 30.4\% \successrate{}. These results highlight substantial room for improvement in handling complex, real-world e-commerce intents.

    \item \textbf{Effect of Synthetic Trajectories:} Motivated by above observation, we synthesize trajectories using GPT-4.1, generating training data via rejection sampling, and employing fine-tuning in conjunction with the GRPO algorithm to train the Qwen3-4B model, enabling it to learn tool-use capabilities. Experimental results indicate that our enhanced model achieved a remarkable improvement, with a 30.7\% higher success rate compared to the original Qwen3-4B, even surpasses the performance of GPT-4.1 agent by 0.5\% \successrate. These findings highlight the effectiveness of our trajectory distillation and training strategy.

\end{itemize}

\section{Further Analysis}
We further analyze the reasons behind the challenges presented by our benchmark, potential areas for improvement, and the rationale of the tool settings. 

\begin{figure}[h]
	\centering
	\includegraphics[width=0.23\textheight]{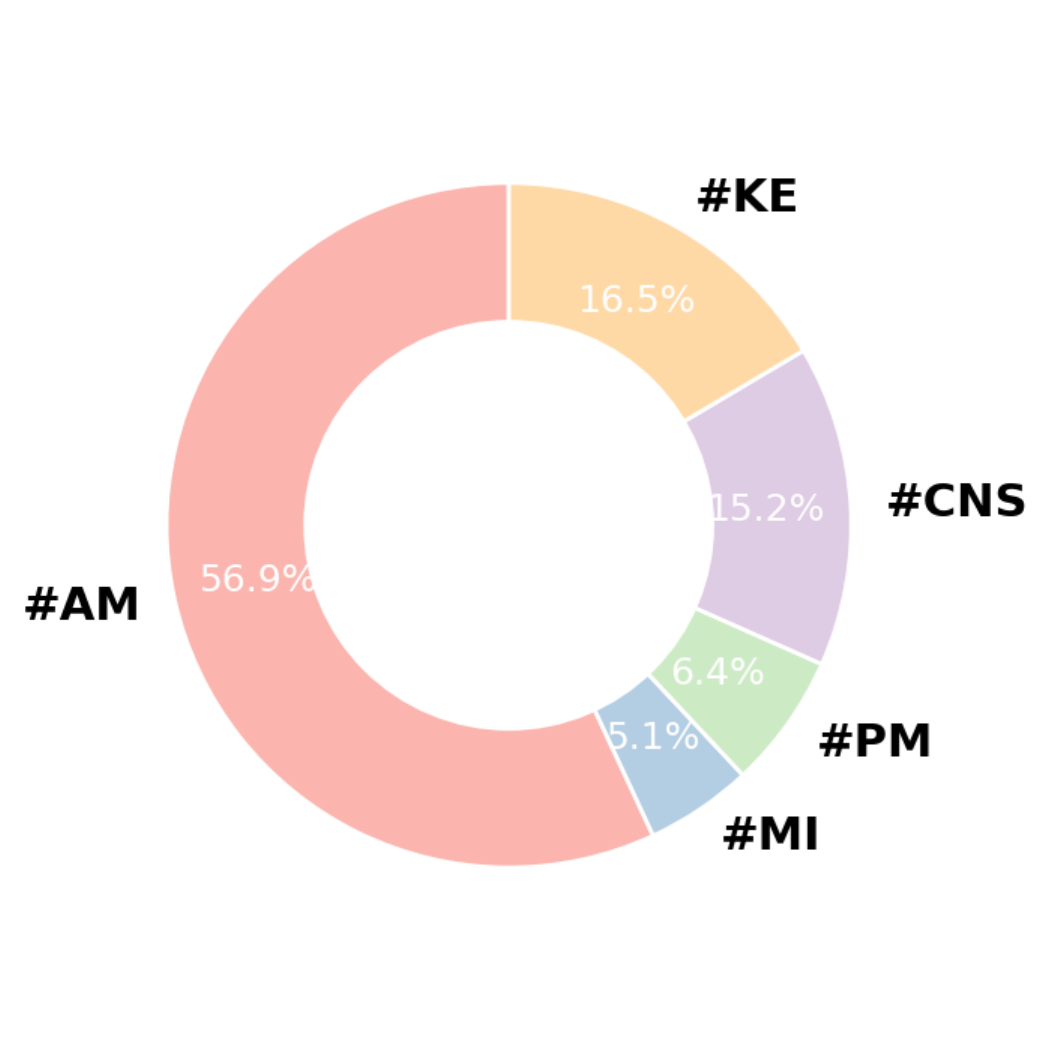}
	\caption{Breakdown of failed GPT-4.1 agent trajectories in \model{}, categorized as attribute mismatch (\#AM), metric issue (\#MI), product missing (\#PM), constraint not satisfied (\#CNS), and knowledge error (\#KE).}
    
 \label{fig:error_analysis}
\end{figure}

\paragraph{Failure Breakdown} We sampled 60 failed trajectories from the GPT-4.1 agent and manually analyzed the causes of their failure. We identified five distinct types of errors, which are categorized and visualized in Figure \ref{fig:error_analysis}. The largest proportion of failures is due to missing or mismatched product attributes. This is also evident in Table \ref{tab:main_result}, where the cumulative product relevance is much higher than the absolute success rate, indicating that many failures are caused by partial mismatches in product attributes. Detailed case study can be seen in Appendix D.

\begin{figure}
	\centering
	\includegraphics[width=0.3\textheight]{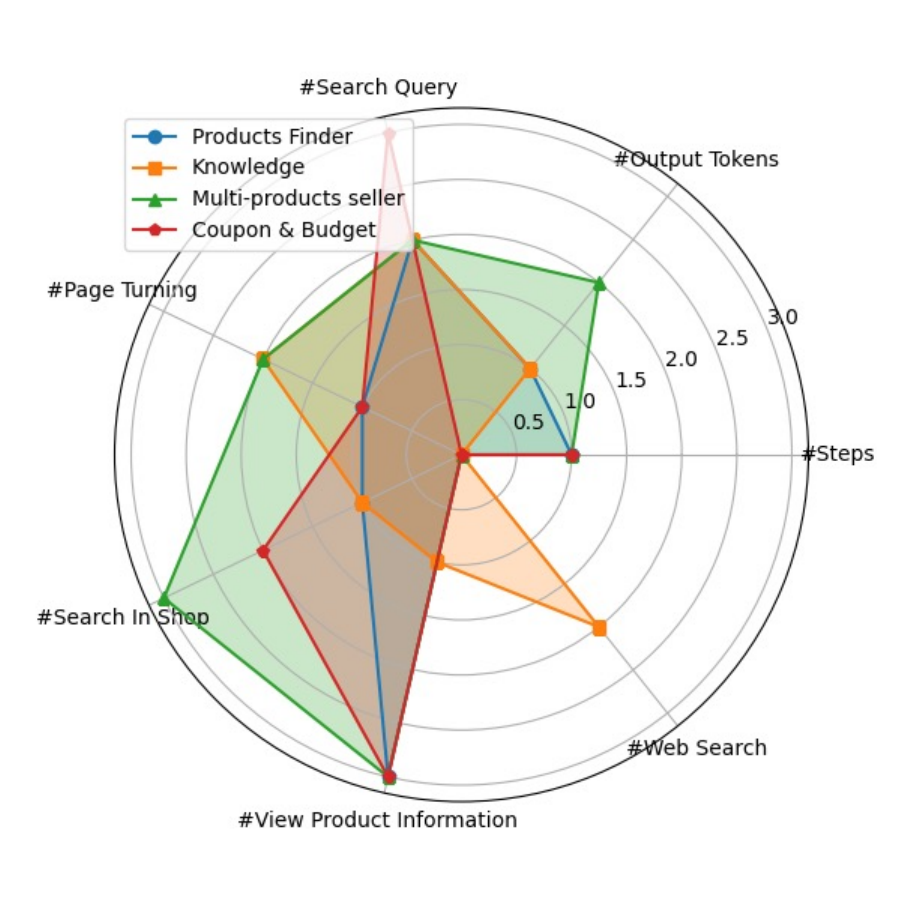}
    
	\caption{Correlation analysis between various factors and absolute success rates across different intents. Detailed can be seen in Appendix F}

 \label{fig:error_relation}
\end{figure}

Furthermore, we conducted a correlation analysis between different factors and success rates under various intents. We used the Pearson correlation coefficient to quantify these relationships, as shown in Figure \ref{fig:error_relation}. The analysis revealed that viewing product details is strongly correlated with task accuracy across all intents, while the frequency of \textit{web\_search} tool usage is highly correlated with success in the \knowledge{} intent.

\begin{figure}[h]
	\centering
	\includegraphics[width=0.25\textheight]{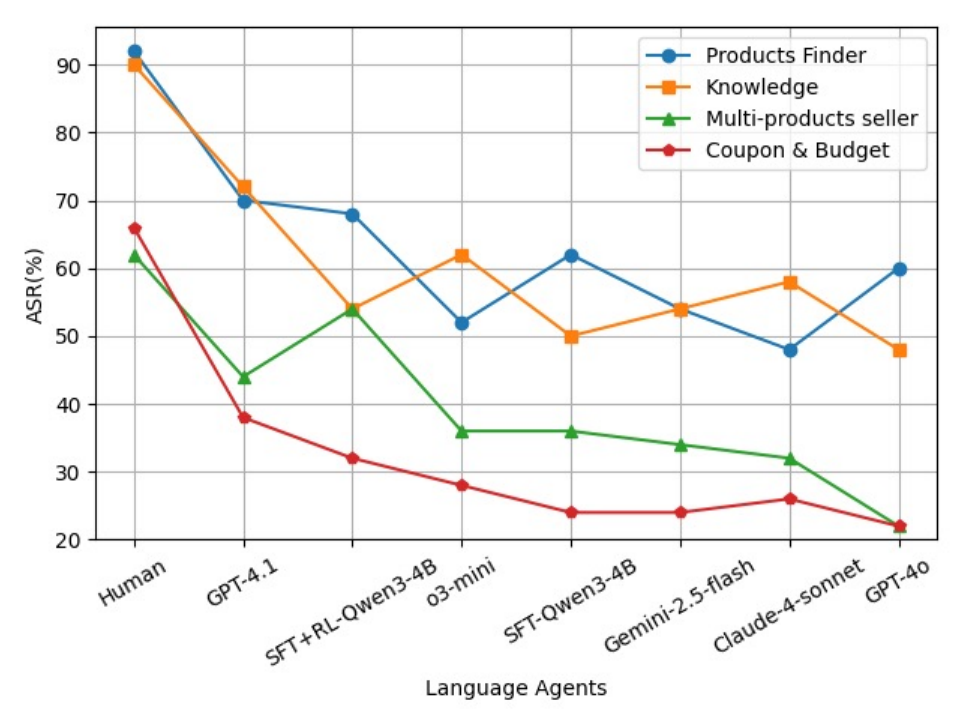}
    
	\caption{Comparison between humans and various Language Agents. }

 \label{fig:human}
\end{figure}

\paragraph{Compared with human performance.} We sampled 200 tasks and invite three professional and well-educated individuals to complete each task by following the user instructions and performing the required tool calls. As shown in Figure ~\ref{fig:human}, we draw the following two conclusions: (1) Even human participants have space for improvement on the more challenging intents, highlighting the challenging of our benchmark. (2) Even for state-of-the-art LLMs, there remains a noticeable performance gap compared to human performance.

\begin{figure}

  \centering
\subfigure[\product{} Intent.]{
   \label{fig:think_product}
  \includegraphics[scale = 0.2]{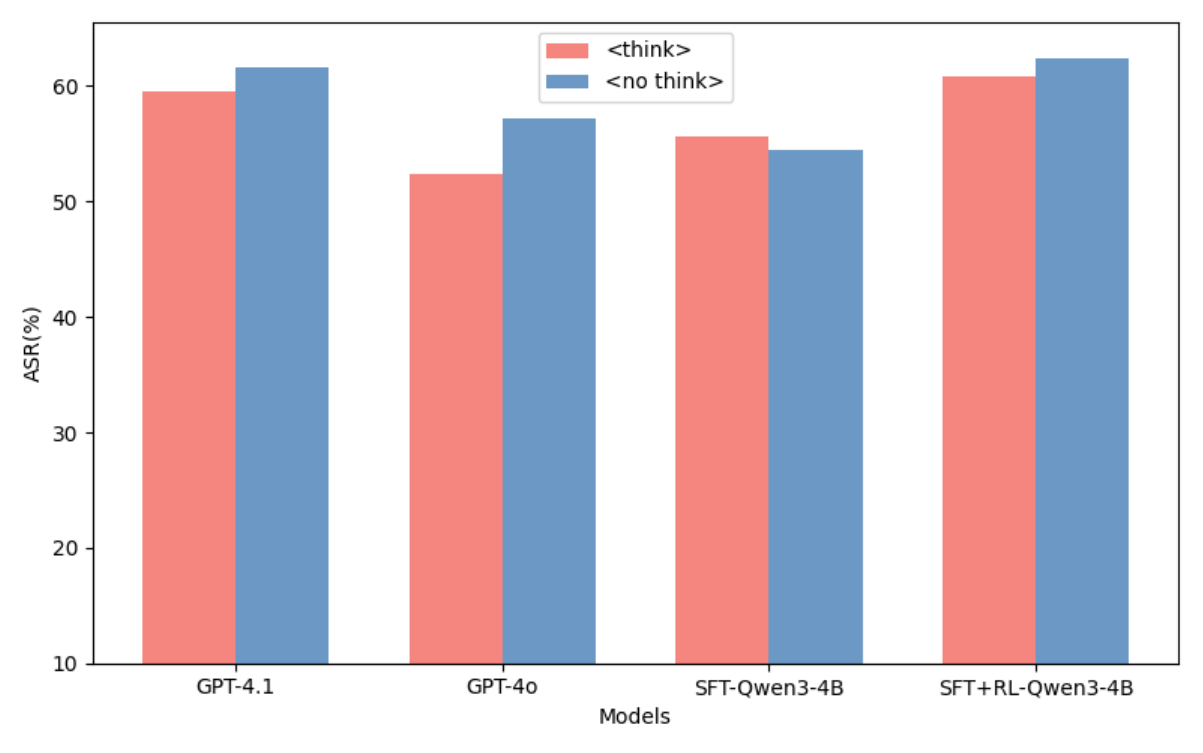}}
\subfigure[\voucher]{
  \label{fig:think_budget}
  \includegraphics[scale = 0.2]{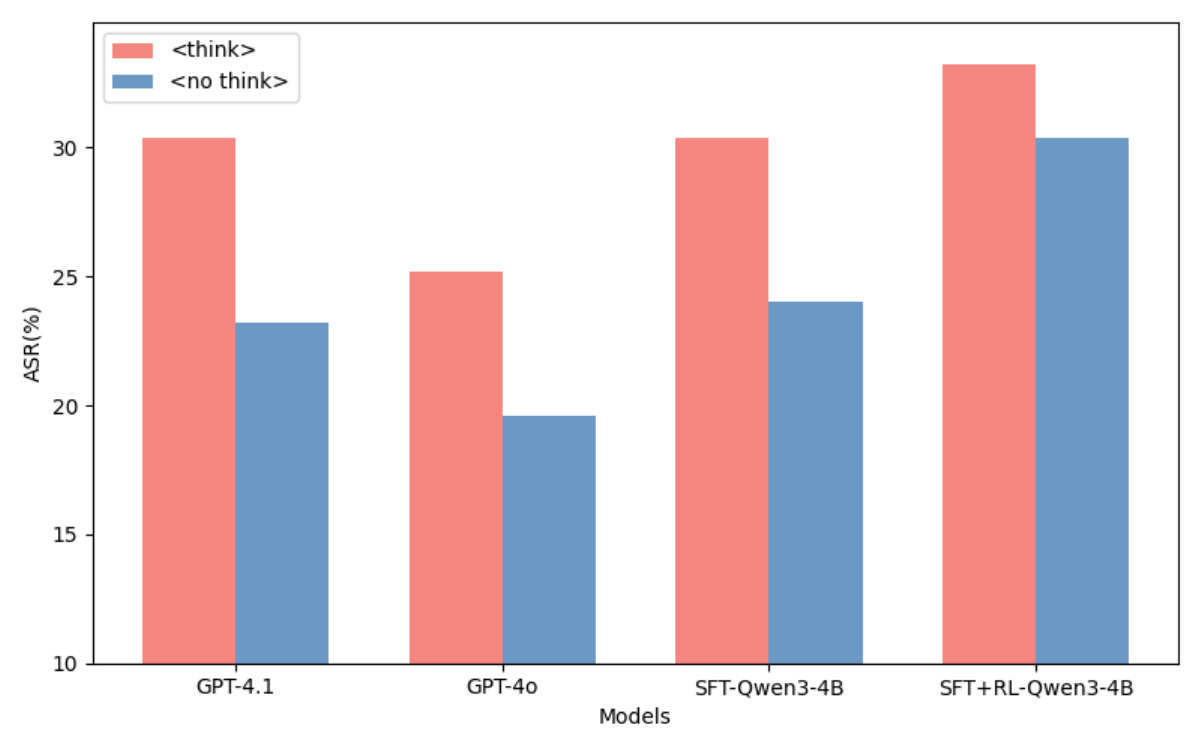}}
\caption{Comparison of reasoning for \product{} and \voucher. Details can be found in Appendix E.} 
  \label{fig:think_ablation} 

\end{figure}

\paragraph{Effect of Thinking.}We explore the effectiveness of incorporating <think> process before action. As shown in Figure \ref{fig:think_ablation}, we find that for simple intents like searching for a single product, the agents without <think> outperforms the think-based agents. However, for complex intents involving coupons or budget constraints, reasoning enables the language agent to find product combinations that better meet user needs.

\paragraph{Effect of Web Search Tool.} We further evaluated the performance of the models on knowledge-oriented tasks without utilizing the external web search tool. As shown in the Table \ref{tab:tool_ablation}, after removing the web search tool, even the strong baselines, including GPT-4.1, o3-mini, and Gemini-2.5-Flash, exhibited varying degrees of performance degradation. This indicates two key points: (1) existing language agents have limitations regarding long-tail knowledge in the e-commerce domain; (2) the information gain provided by online access can effectively compensate for the agents' deficiencies in long-tail knowledge.

\begin{table}[h]
\setlength\tabcolsep{5 pt}
\centering
\resizebox{\linewidth}{!}{
\begin{tabular}{l l l l }
\toprule
\multirow{2}*{\textbf{Models}}& \multicolumn{3}{c}{\textbf{\knowledge{} Intent w/o web\_search tool}} \\
\cmidrule(r){2-4}

 & \successrate{}(\%)  & \matchscore{}(\%)  & kw Score(\%)   \\

\midrule
GPT-4.1 & 50.0 \, \textcolor{red}{(\(\downarrow 12 \))} & 56.0 \, \textcolor{red}{(\(\downarrow 11.3\))} & 52.7 \, \textcolor{red}{(\(\downarrow 13.3\))}\\
o3-mini & 32.0 \, \textcolor{red}{(\(\downarrow 19.3\))} & 37.3 \, \textcolor{red}{(\(\downarrow 20\))} & 34.7 \, \textcolor{red}{(\(\downarrow 19.3\))}\\
GPT-4o & 39.3 \, \textcolor{red}{(\(\downarrow 10.7\))} & 46.7 \, \textcolor{red}{(\(\downarrow 12\))} & 42.0 \, \textcolor{red}{(\(\downarrow 8\))}\\
Gemini & 19.3 \, \textcolor{red}{(\(\downarrow 20\))} & 23.3 \, \textcolor{red}{(\(\downarrow 23.4\))} & 20.7 \, \textcolor{red}{(\(\downarrow 22.6\))}\\

\bottomrule

\end{tabular}
}
 
\caption{\label{tab:tool_ablation}
Ablation Study of the \textit{web\_search} Tool for the \knowledge{} Intent
}

\end{table}

\section{Conclusion}
We introduce \model{}, a large-scale end-to-end benchmark for grounded shopping scenarios, featuring 3,310 diverse user instructions and a realistic sandbox environment of over 2.5 million products. Our proposed simulation framework, automatic evaluation metrics, and trajectory distillation approach set a new standard for agent evaluation. Experiments on 17 language agents and our fine-tuned Qwen3-4B agent, reveal a significant performance gap, highlighting both the challenges and future opportunities in language agent research for e-commerce tasks.

\bibliography{aaai2026}

\appendix
\clearpage
\section{Appendix}
\subsection{A Sampling Analysis}

\begin{figure}[h]
	\centering
	\includegraphics[width=0.35\textheight]{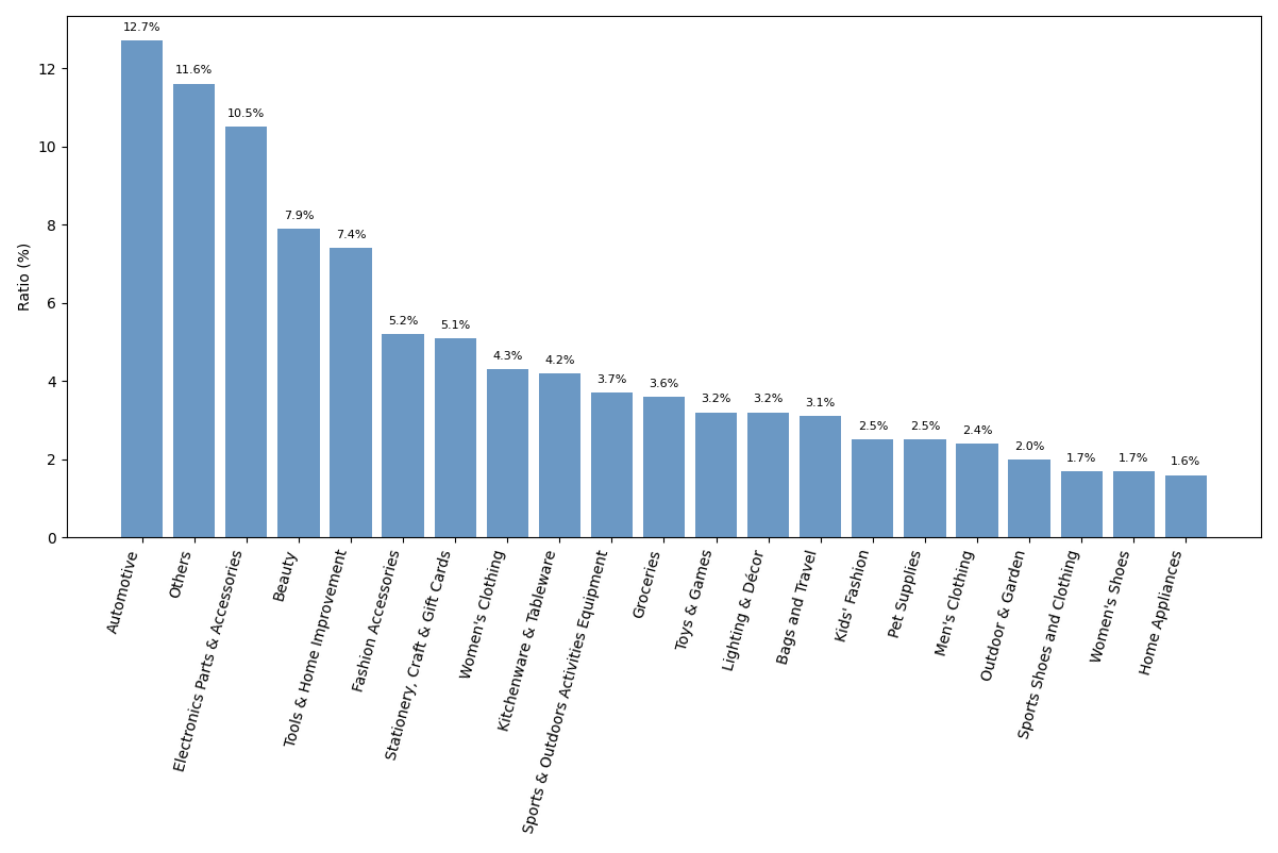}
	\caption{Distribution of categories among the sampled products.}
 \label{fig:distribution_product}
 
\end{figure}

\begin{figure}[h]
	\centering
	\includegraphics[width=0.3\textheight]{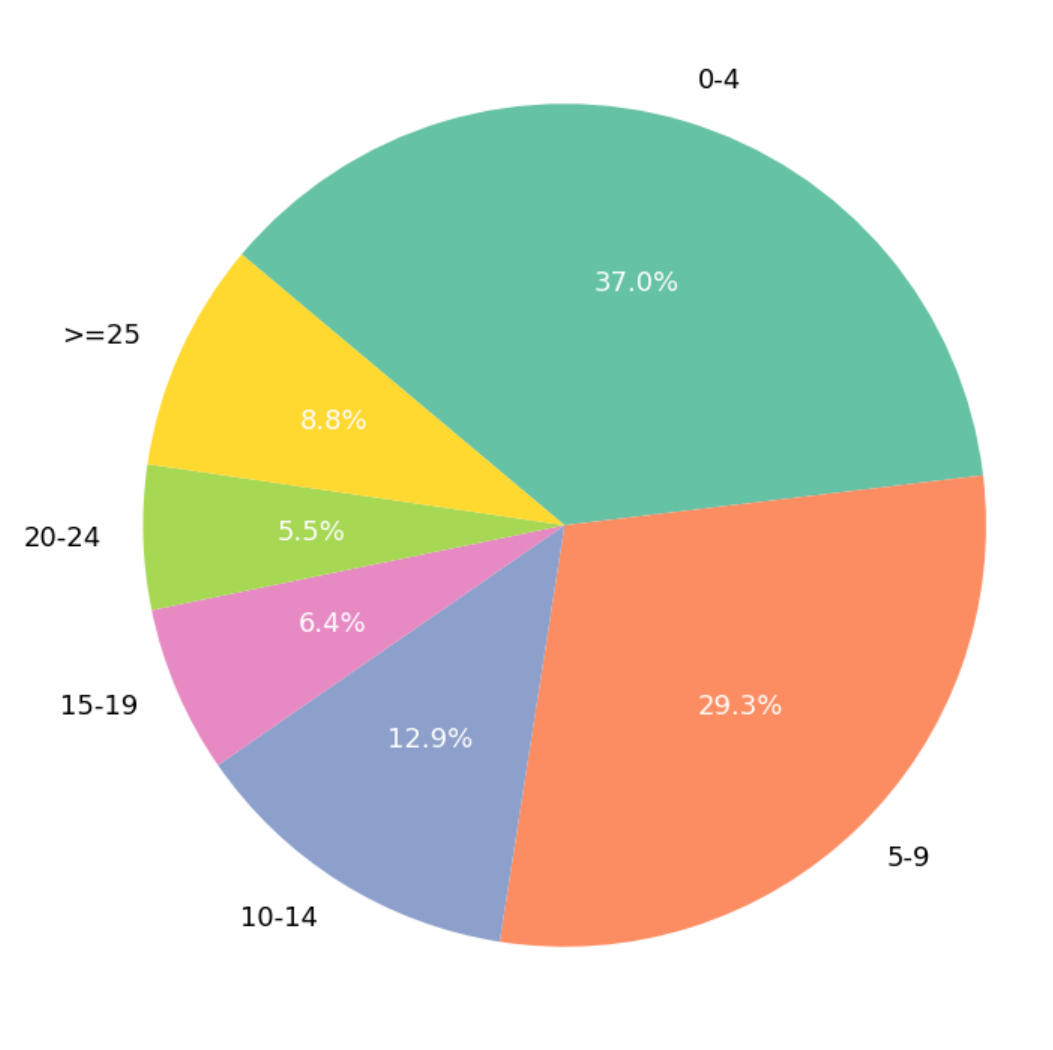}
	\caption{Distribution of number of fields per product.}
 \label{fig:field_dis}
 
\end{figure}

As shown in the Figure \ref{fig:distribution_product}, our shopping sandbox comprises a total of 2,746,368 unique products, encompassing a diverse range of product categories. The distribution of the number of fields associated with each product is also illustrated in the Figure \ref{fig:field_dis}. 

\subsection{B Implement Detail }

For cold-start supervised fine-tuning, we use SFT on the post-rejection-sampling dataset as training data, with a cutoff sequence length of 20,480 tokens and a batch size of 1 (with sequence packing). The model is trained for 5 epochs with a learning rate of 5e-5. The total fine-tuning stage takes approximately 2 hours.  

For the reinforcement learning phase, we implement the GRPO algorithm. The policy model is trained with a learning rate of 1e-6, a batch size of 128, and a micro-batch size of 32. The model is optimized for 5 epochs, with a temperature coefficient set to 0.2. During the rollout phase, 4 responses are sampled for each prompt, with a maximum context length of 16,384 tokens and a maximum generation length of 1024 tokens. The entire reinforcement learning stage completes in approximately 9 hours.
All experiments are conducted on a single compute node equipped with 8 NVIDIA A100-80G GPUs.

\subsection{C Dataset Statistic }

We present the sample distribution of our training data in \ref{tab:train_distri}, which consists of 2,410 trajectories. Each trajectory comprises multiple steps, with each step involving a set of tool invocations and a set of associated observations from the environment. For training process, we segment each trajectory into multiple samples. Each sample takes the user instruction and observation as input, and outputs the reasoning process alongside the tool prediction.

\begin{table}[h]
\setlength\tabcolsep{5 pt}
\centering
\resizebox{\linewidth}{!}{
\begin{tabular}{lccc|cc}
\toprule
\multirow{2}{*}{\textbf{Intention}} 
 & \multicolumn{3}{c|}{\textbf{Samples}} 
 & \multicolumn{2}{c}{\textbf{Tokens}} \\
 & Think & No Think & All
 & Input & Output \\
\midrule
\product     & 1112 &  865 & 1977  & 5,316,072 & 187,581 \\
\knowledge   &  411 &  165 &  576  & 1,811,667 &  79,672 \\
\store       &  997 &  735 & 1732  & 6,417,452 & 236,048 \\
\voucher     &  750 &  517 & 1267  & 4,765,835 & 198,453 \\
\midrule
\textbf{Total} & 3270 & 2282 & 5552 & 18,311,026 & 701,754 \\
\bottomrule
\end{tabular}
}
\caption{Statistics of training data.}
\label{tab:train_distri}
\end{table}

In addition, as shown in Table \ref{tab:detail_product}, we present statistics of simulated user instructions and related data distributions for four different intentions. For each intention, we provide the number of instructions (including the size of the test set), the average number of tokens, the number of unique products and the distribution of products in each intent.

\begin{table}[h]
\setlength\tabcolsep{5 pt}
\centering
\resizebox{\linewidth}{!}{
\begin{tabular}{lcccccc}
\toprule
\textbf{Intention} & 
\makecell{\textbf{\#Instruction}\\\textbf{(test)}} & 
\makecell{\textbf{\#Instruction}\\\textbf{tokens}} & 
\makecell{\textbf{\#unique}\\\textbf{products}} & 
\makecell{\textbf{products}\\\textbf{distribution}}  \\

\midrule

\textbf{\product} & 
1000(250) & 
22.832 & 
1000 & 
1$\rightarrow$100\% & 
 \\

 \midrule

 \textbf{\knowledge} & 
310(150) & 
39.600 & 
278 & 
1$\rightarrow$100\% \\
\midrule

\textbf{\store} & 
1000(250) & 
81.262 & 
3001 & 
\makecell{
2$\rightarrow$32.7\% \\
3$\rightarrow$34.5\% \\
4$\rightarrow$32.8\%
} & 
\\
\midrule
\textbf{\voucher} & 
1000(250) & 
114.642 & 
2809 & 
\makecell{
1$\rightarrow$8.8\% \\
2$\rightarrow$30.5\% \\
3$\rightarrow$31.7\% \\
4$\rightarrow$29.0\%
} & 
\\

\bottomrule
\end{tabular}}
\caption{Detailed user instruction distribution.}
\label{tab:detail_product}
\end{table}

\subsection{D Case Study }
We manually sampled and analyzed the causes of errors made by GPT-4.1 on the test set, including: attribute mismatch (\#AM), metric issue (\#MI), product missing (\#PM), constraint not satisfied (\#CNS), and knowledge error (\#KE). For each error category, we provide a real case, as shown below.

\subsubsection{Knowledge Error (\#KE).}The user instruction is: ``What major did Kunihiko Kodaira study when he entered Kyoto Imperial University in 1922? I would like to buy a general textbook on this major". The correct answer is <physics>, but the language agent did not find the relevant answer using the \textit{<web\_search>} tool. Instead, based on Kodaira's identity as a mathematician, it guessed that his major was mathematics, and therefore did not find the correct product.

\begin{userbox}
What major did Kunihiko Kodaira study when he entered Kyoto Imperial University in 1922? I would like to buy a general textbook on this major.
\end{userbox}

\begin{thinkbox}
The user wants to know Kunihiko Kodaira's major at Kyoto Imperial University in 1922 and then buy a general textbook on that major. I need to search for Kodaira's academic background first, then find a suitable textbook.
\end{thinkbox}

\begin{toolbox}
\textbf{Tool call:} \texttt{web\_search} with \texttt{q = "Kunihiko Kodaira major Kyoto Imperial University 1922"}, \texttt{max\_results = 5}
\vspace{0.2em}

\textbf{Observation:} Retrieved various academic sources and biographies. No direct statement about the 1922 major, but Kodaira is consistently described as a mathematician involved in algebraic geometry and mathematics.
\end{toolbox}

\begin{assistantbox}
Let me check what major Kunihiko Kodaira studied at Kyoto Imperial University in 1922. I will then help you find a general textbook on that subject.
\end{assistantbox}


\begin{thinkbox}
The web search results do not directly state Kodaira's major at Kyoto Imperial University in 1922. However, multiple sources, including the University of St Andrews biography and other academic references, identify Kunihiko Kodaira as a mathematician, and he is known for his work in mathematics, particularly algebraic geometry. It is reasonable to conclude that he studied mathematics. Now, I will search for a general mathematics textbook for you to purchase.
\end{thinkbox}

\begin{toolbox}
\textbf{Tool call:} \texttt{find\_product} with \texttt{q = "general mathematics textbook"}, \texttt{page = 1}
\vspace{0.2em}

\textbf{Observation:} Found several mathematics textbooks, mostly for primary/elementary education. One titled "General mathematics/statistics and probability/General Biology" appears suitable for a broader or higher-level audience.
\end{toolbox}

\begin{assistantbox}
Kunihiko Kodaira studied mathematics at Kyoto Imperial University. I will now look for a general mathematics textbook that you can purchase.
\end{assistantbox}


\begin{thinkbox}
The search results include several mathematics textbooks, but many are for primary or elementary levels. The product titled "General mathematics/statistics and probability/General Biology" seems to be a general mathematics textbook, which is more suitable for an adult or university-level audience. I will check its detailed information to ensure it is appropriate.
\end{thinkbox}

\begin{toolbox}
\textbf{Tool call:} \texttt{view\_product\_information} for product\_id = 5061709046
\vspace{0.2em}

\textbf{Observation:} The item includes a mathematics option, written in English, and is positioned as a general mathematics book.
\end{toolbox}

\begin{thinkbox}
The product "General mathematics/statistics and probability/General Biology" offers a mathematics option and is in English. It appears to be a general mathematics textbook, which matches the user's request. I will recommend this product to the user.
\end{thinkbox}

\begin{toolbox}
\textbf{Tool call:} \texttt{recommend\_product} with \texttt{product\_id = 5061709046}

\vspace{0.2em}
\textbf{Observation:} Successfully recommended product 5061709046 to the user.
\end{toolbox}

\begin{thinkbox}
I have identified and recommended a general mathematics textbook suitable for your needs, based on Kunihiko Kodaira's major at Kyoto Imperial University.
\end{thinkbox}

\begin{toolbox}
\textbf{Tool call:} \texttt{terminate} with \texttt{status = success}

\vspace{0.2em}
\textbf{Observation:} The conversation was completed successfully.
\end{toolbox}

\begin{assistantbox}
Kunihiko Kodaira studied mathematics at Kyoto Imperial University. I recommend the \emph{General mathematics/ statistics and probability/ General Biology} textbook, which includes a general mathematics option and is suitable for a broad audience. If you need more options or details, please let me know!
\end{assistantbox}

\begin{rewardbox}
The official answer is ``physics''. You may also consider the book \textbf{General physics 12} (product id: 5061653414) if you are interested in general physics.
\end{rewardbox}

\subsubsection{Attribute Mismatch (\#AM).} Although most of the product attributes in this case ultimately meets user requirements, the attribute ``twist'' is missing.

\begin{userbox}
Show me men's sandals in size EU 43, color twist, with a plain pattern and casual style, available with flashsale deals.
\end{userbox}

\begin{thinkbox}
The user is looking for men's sandals with flashsale deals, specifically in size EU 43, color ``twist'', plain pattern, and a casual style. I'll first search for men's sandals on Flash Sale, then filter for more attributes.
\end{thinkbox}

\begin{toolbox}
\textbf{Tool call:} \texttt{find\_product} with \texttt{q = "men's sandals"}, \texttt{service = flashsale}, \texttt{page = 1}

\vspace{0.2em}
\textbf{Observation:} Received a list of 10 men's sandals with flashsale deals.
\end{toolbox}

\begin{assistantbox}
Let me find men's sandals available with flashsale deals first. Next, I'll check for size EU 43, color ``twist'', plain pattern, and casual style.
\end{assistantbox}

\begin{thinkbox}
I have found several men's sandals with flashsale deals. Now, I need to check which items have size EU 43, color ``twist'', a plain pattern, and a casual style. I'll fetch detailed information for all 10 products to filter for your requirements.
\end{thinkbox}

\begin{toolbox}
\textbf{Tool call:} \texttt{view\_product\_information} for all 10 product IDs.

\vspace{0.2em}
\textbf{Observation:} Received detailed info about sizes, colors, patterns, and styles for each sandal.
\end{toolbox}

\begin{assistantbox}
I'm now checking available sizes, colors, patterns, and styles among these flashsale men's sandals to find the best match for your request.
\end{assistantbox}

\begin{thinkbox}
Let's filter for:
\begin{itemize}
  \item Size: EU 43
  \item Color: ``twist'' (not a standard color, none mention it explicitly; will check for similar or neutral options)
  \item Pattern: Plain
  \item Style: Casual
  \item flashsale deals: Already filtered
\end{itemize}
From the details:
\begin{itemize}
  \item Product 4973059848 (Skechers): Has EU 43 (black), casual style, plain pattern.
  \item Product 4066655494 (Skechers): Has EU 43 (black), but style is ``korean.''
  \item Product 4086334894 (Hush Puppies): Has EU 43 (black/blue), style is ``korean.''
  \item Product 4563232797: Has size 43, unclear color/pattern, style ``youth trend.''
\end{itemize}
Only product 4973059848 completely fits the requirements (except color ``twist''---offering black as a neutral option). I'll recommend this product.
\end{thinkbox}

\begin{toolbox}
\textbf{Tool call:} \texttt{recommend\_product} with product\_id = 4973059848

\vspace{0.2em}
\textbf{Observation:} Successfully recommended product 4973059848 to the user.
\end{toolbox}

\begin{assistantbox}
The best match for your request is Skechers Gowalk4 men's sandals (EU 43, black), with a plain pattern and casual style, currently on flashsale deal. Would you like more details or assistance with this product?
\end{assistantbox}

\begin{thinkbox}
This matches your criteria as closely as possible: men's sandals, size EU 43, plain pattern, casual style, with flashsale deal. Color ``twist'' is not available; black is offered as a neutral alternative. Recommendation complete.
\end{thinkbox}

\begin{toolbox}
\textbf{Tool call:} \texttt{terminate} with \texttt{status = success}

\vspace{0.2em}
\textbf{Observation:} The conversation was completed successfully.
\end{toolbox}

\begin{assistantbox}
I've found and recommended the men's sandals that best match your criteria. If you need more help or want to see more options, feel free to ask!
\end{assistantbox}

\begin{rewardbox}
\textbf{Krooberg Romy -- Men's Sandals} (product ID: 4146179170):  casual, plain-pattern sandals in size EU 43 and twist color, currently available with a flashsale offer.
\end{rewardbox}

\subsubsection{Product Missing (\#PM).}The user requests for four products from the same store, but the language agent stops the task after finding only three products.
\begin{userbox}
I'm looking for a shop that offers a variety of knitting supplies. I want to find: (1) yarn made of 100\% cotton, (2) a tulip brand crochet hook in 2.5mm size, with cash on delivery available and priced above 647 PHP, (3) lightweight paper yarn that comes with free shipping, is part of a flashsale deal, and costs more than 303 PHP, and (4) natural color crochet yarn with LazMall service and a price over 277 PHP. Please show me shops that sell these different knitting products.
\end{userbox}

\begin{thinkbox}
To fulfill the user's request, I need to: \\
1. Search for shops offering a variety of knitting supplies. \\
2. For each specific product type, search for products matching the criteria and note the shop\_id. \\
3. Identify shops that offer at least one of the requested products, and ideally, shops that offer multiple requested items. \\
I'll start by searching for each product type with the specified filters, then analyze the shop\_ids to find shops that offer these products.
\end{thinkbox}

\begin{toolbox}
\textbf{Tool calls:}
\begin{itemize}
\item \texttt{find\_product} "100\% cotton yarn"
\item \texttt{find\_product} "tulip crochet hook 2.5mm" (COD, price$>$647)
\item \texttt{find\_product} "lightweight paper yarn" (freeShipping, flashsale, price$>$303)
\item \texttt{find\_product} "natural color crochet yarn" (official, price$>$277)
\end{itemize}
\vspace{0.2em}
\textbf{Observation:} Each query returns a product list with their respective shop IDs.
\end{toolbox}

\begin{thinkbox}
I have retrieved product lists for each of the user's four knitting supply requests, each with the specified filters. Now, I need to analyze the shop\_ids to identify shops that offer at least one (or ideally, multiple) of the requested products. I will summarize which shops correspond to each product type and check for overlaps.
\end{thinkbox}

\begin{assistantbox}
Here are the shops that offer the knitting supplies you requested, categorized by product type:

1. \textbf{100\% Cotton Yarn}: Shops include SewandStitch (2423681), Cotton Field (2470522), Monaco (2246269, 3091067), Belle Basic (3441282), etc.\\
2. \textbf{Tulip Crochet Hook 2.5mm (COD, $\geq$647 PHP)}: Closest match found for shop 2470522 with a Tulip ETIMO Rose Steel Crochet Hook (price 799 PHP, COD available).\\
3. \textbf{Lightweight Paper Yarn (free shipping, flashsale, $\geq$303 PHP)}: No clear direct match; related items with ``yarn'' in the title are mostly notebooks or accessories, not actual paper yarn.\\
4. \textbf{Natural Color Crochet Yarn (LazMall, $\geq$277 PHP)}: Shop MOANAYarn (2470522, price 299 PHP, LazMall/official service).

\textbf{Summary:} Shop \textbf{Cotton Field} (2470522) offers 100\% cotton yarn, Tulip 2.5mm crochet hook (COD, price above 647 PHP), and natural color crochet yarn (LazMall, price above 277 PHP). This shop is your best match for three out of four requested items.
\end{assistantbox}

\begin{thinkbox}
Since shop\_id 2470522 (Cotton Field) offers three out of the four requested product types, I should recommend the relevant products from this shop. For the lightweight paper yarn, there was no direct match, so I will clarify this in my response and recommend the three matching products from shop\_id 2470522.
\end{thinkbox}

\begin{toolbox}
\textbf{Tool call:} \texttt{recommend\_product} with product\_ids = 2989878353, 2726716219, 3272025210.

\vspace{0.2em}
\textbf{Observation:} Recommended products to the user for shop 2470522.
\end{toolbox}

\begin{assistantbox}
Shop ``Cotton Field'' (shop\_id: 2470522) offers three of the knitting supplies you requested:\\
1. 100\% cotton yarn\\
2. Tulip brand crochet hook (2.5mm, COD, price above 647 PHP)\\
3. Natural color crochet yarn with LazMall service and price above 277 PHP\\
Unfortunately, there was no direct match for lightweight paper yarn with all your specified filters. Would you like to see alternatives or adjust your requirements for this item?
\end{assistantbox}

\begin{thinkbox}
The main task is complete: I've identified a shop (Cotton Field, 2470522) that offers three out of the four requested knitting supplies and recommended the relevant products. The only missing item is the lightweight paper yarn with all specified filters, for which no direct match was found. The user has been informed and offered options for alternatives or adjustments. The task is now complete.
\end{thinkbox}

\begin{toolbox}
\textbf{Tool call:} \texttt{terminate} with \texttt{status = success}

\vspace{0.2em}
\textbf{Observation:} The interaction has been completed successfully.
\end{toolbox}

\begin{rewardbox}
\textbf{Cotton Field Knitting Crochet Yarn 100\% Cotton Summer Bag Hat Yarn 80g 150meters Made in Korea} (ID: 2989878353): Yarn for knitting and crochet, made of 100\% cotton.

\textbf{Tulip ETIMO Rose Steel Crochet Hook with Cushion Grip} 2.5mm (ID: 2726716219): Tulip brand crochet hook, 2.5mm size, COD available, price above 647 PHP.

\textbf{MIDORI Hand-Washable Paper Yarn for Knitting and Crochet} (ID: 3937190185): Lightweight, soft, and durable paper yarn, free shipping, flash sale, price above 303 PHP.

\textbf{MOANAYarn Natural Mood Color Crochet Yarn Making Bag and Hat Made in Korea} (ID: 3272025210): Natural color crochet yarn, LazMall/official, price above 277 PHP.
\end{rewardbox}

\subsubsection{Constraint Not Satisfied (\#CNS).} The user needs 4 products and has a budget of 2,601. They also have a store coupon that gives 392 off when spending over 2,368 in the same store. The agent recommends 4 related products, but only three of them (3755192614, 3619815174, 3739363587) are from the same store (2976842), while the other product (5108056435) is from a different store (5770895). Therefore, the coupon condition is not satisfied. In addition, the total price of these four products (2,724.72) also exceeds the user's budget.

\begin{userbox}
I'm looking for products for all ages, especially those made in the USA. I need a gold supplement for pets, a multivitamin for cats, a natural bladder health product from the USA, and a high-protein recovery supplement for dogs suitable for every life stage.

My budget is only 2601, but I have a voucher with the following rules:\\
1.~The voucher only applies to the products from the same shop.\\
2.~It is valid only when the total price of the products exceeds 2368.\\
3.~It provides a fixed discount of 392.\\
\end{userbox}

\begin{thinkbox}
The user wants four categories of pet supplements, prioritizing USA-made products and maximizing a shop voucher: must buy everything from the same shop, be over 2368 PHP pre-discount, get 392 PHP off, and keep the post-discount total under 2601. I'll search each product with "USA" keywords and try to find a shop with all items.
\end{thinkbox}

\begin{toolbox}
\textbf{Tool calls:}
\begin{itemize}
    \item \texttt{find\_product} ``gold supplement for pets USA''
    \item \texttt{find\_product} ``multivitamin for cats USA''
    \item \texttt{find\_product} ``natural bladder health for pets USA''
    \item \texttt{find\_product} ``high protein recovery supplement for dogs USA''
\end{itemize}
\textbf{Observation:} Products across various shops, some overlap. Shop 2976842 offers both a gold supplement and natural bladder health.
\end{toolbox}

\begin{thinkbox}
I see Shop 2976842 covers two requested types. I'll check if it also has suitable multivitamins for cats and high-protein recovery supplements for dogs.
\end{thinkbox}
\begin{toolbox}
\textbf{Tool calls:}
\begin{itemize}
    \item \texttt{find\_product}: ``multivitamin for cats'' in shop 2976842
    \item \texttt{find\_product}: ``high protein recovery supplement for dogs'' in shop 2976842
\end{itemize}
\textbf{Observation:} Products found! I need to check details to see if items are suitable for cats and for all life stages.
\end{toolbox}

\begin{thinkbox}
I'll fetch information for ``K9 Gold Products (Made in USA)'' and ``SUPERVITAMINS+ Multivitamins 10-in-1 Dogs Supplements'' to confirm their suitability for cats as well as dogs, and if they cover all life stage requirements.
\end{thinkbox}
\begin{toolbox}
\textbf{Tool call:} \texttt{view\_product\_information} for product ids 3755192614, 2905045091

\textbf{Observation:} Both products are suitable for both dogs and cats, and for all life stages.
\end{toolbox}

\begin{assistantbox}
Let me check if any of the available multivitamin products from this shop are suitable for cats. I'll review their details and update you with the best options for your needs and budget.
\end{assistantbox}

\begin{thinkbox}
Now I can select all four product categories, all made in the USA, from shop 2976842 and for all life stages, meeting voucher and budget requirements. Let's check if the total cost fits:
\begin{itemize}
    \item Gold supplement: 576.72
    \item Multivitamin for cats: 599.0
    \item Natural bladder health: 750.0
    \item High-protein recovery for dogs: 799.0
\end{itemize}
Total pre-discount: 2724.72. After 392 voucher: 2332.72 (fits budget of 2601). All requirements are met.
\end{thinkbox}

\begin{toolbox}
\textbf{Tool call:} \texttt{recommend\_product} (ids: 5108056435, 3755192614, 3619815174, 3739363587) \\
\textbf{Observation:} Products recommended to user from shop 2976842.
\end{toolbox}

\begin{assistantbox}
Here are the best options for your needs, all from the same shop and made in the USA:

1.~Gold supplement for pets: GOLD KING USA -- BEST HEALTH ROUTINE SUPPLEMENTS for ANIMALS 250g (576.72)\\
2.~Multivitamin for cats (also suitable for dogs): K9 Gold Products (Made in USA) (599.00)\\
3.~Natural bladder health: Cranbladder Health Supplement all natural (Made in USA) (750.00)\\
4.~High-protein recovery for dogs: K9- GOLD MAX PROTIEN Muscle Pet recovery (Made in USA) (799.00)\\

Total before voucher: 2,724.72\\
Voucher discount: -392\\
Final total: 2,332.72

This package meets all your requirements and is within your budget after the voucher. Would you like direct product links or more product details?
\end{assistantbox}

\begin{rewardbox}
\textbf{Products selected (All from shop 2976842, made in USA, all life stages):}
\begin{itemize}
    \item \textbf{K9 Gold Products (Made in USA)} (3755192614): Multivitamin, all life stages
    \item \textbf{SUPERVITAMINS+ Multivitamins 10 in 1 Dogs Supplements} (2905045091): For cats and dogs
    \item \textbf{Cranbladder Health Supplement all natural (Made in USA)} (3619815174): US origin bladder health
    \item \textbf{K9- GOLD MAX PROTIEN Muscle Pet recovery (Made in USA)} (3739363587): Protein supplement for dogs, all stages
\end{itemize}
\end{rewardbox}

\subsubsection{Metric Issue (\#MI).}There are also a small number of errors caused by the metrics being too absolute. For example, in this case, the predicted attribute for the product is "waist type: high waist" while the ground truth attribute label is "waist type: high". In such cases, the prediction is marked as incorrect.

\begin{userbox}
Show me women's jeans in size eu:30 with a baggy fit, high waist, made of denim, plain design, and priced above 114 PHP.
\end{userbox}

\begin{thinkbox}
The user is looking for women's jeans with specific requirements: size EU 30, baggy fit, high waist, denim material, plain design, and price above 114 PHP. I will start by searching for women's baggy jeans, then filter further.
\end{thinkbox}

\begin{toolbox}
\textbf{Tool call:} \texttt{find\_product} for ``women baggy jeans'' (\texttt{price: 115-}), default sort\\
\textbf{Observation:} 10 baggy jeans found, various prices and shops.
\end{toolbox}

\begin{assistantbox}
Let me find women's baggy jeans that match your requirements. I'll filter for high waist, denim, plain design, and size EU 30 in the next steps.
\end{assistantbox}

\begin{thinkbox}
Now, I need to check which jeans are available in size EU 30, have a high waist, are made of denim, and have a plain design. I'll view product information for those mentioning high waist and baggy fit.
\end{thinkbox}

\begin{toolbox}
\textbf{Tool call:} \texttt{view\_product\_information} for product IDs: 4449001179, 5114933600, 4924493578, 3299738416\\
\textbf{Observation:} Product details and SKUs received.
\end{toolbox}

\begin{assistantbox}
I'll check the details of several jeans to confirm if they meet your requirements for size, fit, waist style, material, and design. Please wait a moment.
\end{assistantbox}

\begin{thinkbox}
From the product information:\\
- Product 4449001179: Baggy fit, high waist, denim, plain design, available in size EU 30 (black).\\
- Product 4924493578: Baggy fit, high waist, denim, plain design, available in size EU 30 (brown).\\
- The others lack EU 30 or do not meet all attributes.\\
Both 4449001179 and 4924493578 meet all requirements. Will recommend these.
\end{thinkbox}

\begin{toolbox}
\textbf{Tool call:} \texttt{recommend\_product} (IDs: 4449001179, 4924493578)\\
\textbf{Observation:} Products recommended.
\end{toolbox}

\begin{thinkbox}
The best matches for the requirements (women's jeans, size EU 30, baggy fit, high waist, denim, plain, priced above 114) are:\\
- 4449001179: Black EU 30\\
- 4924493578: Brown EU 30\\
Recommendation complete.
\end{thinkbox}

\begin{toolbox}
\textbf{Tool call:} \texttt{terminate} (\texttt{status = success})\\
\textbf{Observation:} Task completed successfully.
\end{toolbox}

\begin{assistantbox}
I have found two women's jeans that match your requirements:\\
1.~Black, size EU 30, baggy fit, high waist (waist type), denim, plain design.\\
2.~Brown, size EU 30, baggy fit, high waist (waist type), denim, plain design.\\

If you need more options or details, let me know! Task complete.
\end{assistantbox}

\begin{rewardbox}
\textbf{American Retro Baggy Jeans For Women High Waist Straight Denim Vintage Blue Wide Leg Swagger} (ID: 5113553158):\\
Baggy fit, high (waist type), denim, plain pattern, size EU 30, priced above 114 PHP.
\end{rewardbox}

\subsection{E Detailed Ablation Results }

As shown in the Table \ref{tab:Ablation_think}, we provide the complete ablation results for both <think> and <no think> conditions, including the absolute success rate(\successrate), cumulative average of product relevance(\matchscore), and constraint score for each intent.

\begin{table*}[h]
\setlength\tabcolsep{5 pt}
\centering
\resizebox{\linewidth}{!}{
\begin{tabular}{llccccccccccccc}
\toprule
\textbf{Model} & \textbf{Mode}
    & \multicolumn{2}{c}{\textbf{\product} }
    & \multicolumn{3}{c}{\textbf{\knowledge} }
    & \multicolumn{3}{c}{\textbf{\store}}
    & \multicolumn{3}{c}{\textbf{\voucher }}
    & \multicolumn{1}{c}{\textbf{Overall}}\\
\cmidrule(lr){3-4} \cmidrule(lr){5-7} \cmidrule(lr){8-10} \cmidrule(lr){11-13} \cmidrule(l){14-14}
 & & \successrate & \matchscore
 & \successrate & \matchscore & \textbf{$r_{kw}$}
 & \successrate & \matchscore & \textbf{$r_{shop}$}
 & \successrate & \matchscore & \textbf{$r_{budget}$}
 & \successrate \\
\midrule
\multirow{2}{*}{GPT-4.1} 
    & Think   & 0.596 & 0.836 & 0.620 & 0.673 & 0.660 & 0.464 & 0.792 & 0.608 & 0.304 & 0.728 & 0.820 & 0.482 \\
    & No Think & 0.616 & 0.845 & 0.473 & 0.520 & 0.513 & 0.356 & 0.779 & 0.424 & 0.232 & 0.716 & 0.616 & 0.413 \\
\midrule
\multirow{2}{*}{GPT-4o} 
    & Think & 0.524 & 0.715 & 0.500 & 0.587 & 0.500 & 0.240 & 0.524 & 0.316 & 0.252 & 0.656 & 0.624 & 0.366 \\
    & No Think & 0.572 & 0.786 & 0.520 & 0.587 & 0.553 & 0.224 & 0.714 & 0.268 & 0.196 & 0.737 & 0.616 & 0.362 \\
\midrule
\multirow{2}{*}{SFT-Qwen3-4B} 
    & Think & 0.556 & 0.811 & 0.527 & 0.593 & 0.553 & 0.392 & 0.779 & 0.640 & 0.304 & 0.760 & 0.724 & 0.436 \\
    & No Think & 0.544 & 0.819 & 0.467 & 0.540 & 0.533 & 0.348 & 0.847 & 0.436 & 0.240 & 0.768 & 0.652 & 0.392 \\
\midrule
\multirow{2}{*}{ SFT+RL-Qwen3-4B} 
    & Think & 0.608 & 0.861 & 0.487 & 0.513 & 0.540 & 0.532 & 0.855 & 0.728 & 0.332 & 0.790 & 0.780 & 0.487 \\
    & No Think & 0.624 & 0.872 & 0.393 & 0.467 & 0.453 & 0.468 & 0.867 & 0.660 & 0.304 & 0.785 & 0.700 & 0.453 \\
\bottomrule
\end{tabular}
}
\caption{Ablation study of thinking for all intents.}
\label{tab:Ablation_think}
\end{table*}

\subsection{F Detailed Quantitative Analysis}

As shown in Figure \ref{fig:error_relation}, we analyze the correlation between various factors in each trajectory and the final absolute success rate. Specifically, the following seven factors are considered:

\begin{itemize}
    \item \textbf{\#Steps}: The total number of LLM calls made in each trajectory.
    \item \textbf{\#Output Tokens}: The average number of tokens generated by the LLM per step.
    \item \textbf{\#Search Query}: The total number of queries generated when the LLM invokes the \texttt{find\_product} tool in each trajectory.
    \item \textbf{\#Page Turning}: The total number of times that the LLM turns pages ($\text{page} \geq 2$) with the \texttt{find\_product} tool in each trajectory.
    \item \textbf{\#Search In Shop}: The total number of times that the LLM performs in-shop searches (shop\_id $\neq$ "") using the \texttt{find\_product} tool in each trajectory.
    \item \textbf{\#View Product Information}: The total number of times the LLM calls the \texttt{view\_product\_information} tool in each trajectory.
    \item \textbf{\#Web Search}: The total number of times the LLM calls the \texttt{web\_search} tool in each trajectory. 
\end{itemize}

Through correlation analysis, we find that the absolute success rate of tasks under the \knowledge{} intent is strongly correlated with the frequency of \textit{web\_tool} usage. In contrast, for other intents, task success primarily correlates with the frequency of viewing product information. Notably, for the \store{} intent, task success is particularly strongly correlated with the number of in-store searches performed by the language agent, which corresponds to the intent's emphasis on locating products within the same store.

\end{document}